# High Precision In-Pipe Robot Localization with Reciprocal Sensor Fusion – 19516

Dapeng Zhao *, William Whittaker *
* Carnegie Mellon University

**ABSTRACT**

Manual measurement of U-235 deposits in uranium enrichment piping is a costly, time consuming, and labor-intensive process that is a well-known cost and schedule driver. Autonomous, robotic innovation enabled by this research is revolutionizing the measurement speed, quality and safety of this important operation. The huge advantage is the robot's ability to measure from inside the pipes. The upside is sensing the geometry, appearance and radiometry directly. The downside is the inability to know precise, absolute position of the measurements in very long pipe runs. This paper develops the unprecedented localization required for this purpose.

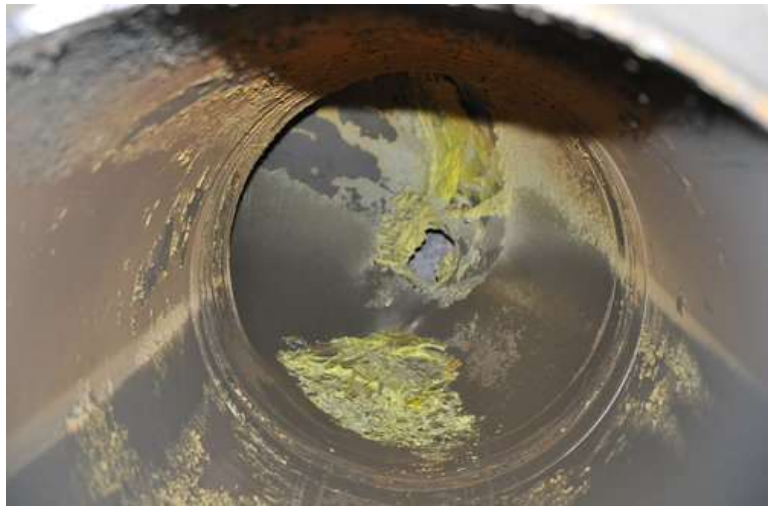

Fig. 1: Uranium holdup deposit in process piping

This paper presents the precise localization method designed for in-pipe radiation measurement robots. Unlike other in-pipe localization methods, this approach achieves millimeter-level accuracy in hundred-foot runs and does everything on one robot without relying on tether cable or communicating devices out of pipe, which is an important feature for nuclear in-pipe robot. It overcomes challenges usually encountered by in-pipe localization such as long travelling distance and operation time, narrow view angle, and featureless environment.

The robot is equipped with encoders embedded in tracks which can record travelled distance, but encoder odometry drifts unacceptably due to slip and non-linearity. The robot also incorporates a laser rangefinder measuring the absolute distance, but rangefinder in long pipes has many misleading false measurements that are difficult to filter. The innovation here is to use one to filter/calibrate the other so that data used from the two sensors improve each other iteratively and reciprocally. The survey measurements accuracy of this localization method is proven by experiments comparing to ground truth and "Zippering error" experiments that compare measurements to fixed features in pipes.

## INTRODUCTION

**Background**

Vast amounts of U-235 remain in miles of piping that once enriched America's uranium. These immense facilities are now defunct and decommissioning is underway. An immense driver of decommissioning schedule and budget is a requirement to determine the exact grams of U-235 in every foot of that pipe before demolition.

To date, human workers in protective clothing have manually deployed detectors from the outside of these pipes to observe radiation emanating from the U-235 inside the pipes. This incurs operational disadvantages of clearing around pipes for access, hazards of elevated work, rad exposure, and manual data transcription. The technical disadvantages include faint signal from attenuation through pipe walls, inability to directly view the deposits, and inability to position a detector on the pipe's center-line.

Since decommissioning involves cutting pipes, the unique opportunity is to robotically deploy detectors from the inside rather than the outside to measure per-foot quantities of U-235. Compared to the current manual method, in-pipe robotic measurement achieves superior speed, accuracy and certainty. It does not require demolition for clear access around pipes. It provides video and geometric record of deposit acquired from inside pipes and precludes significant elevated human work.

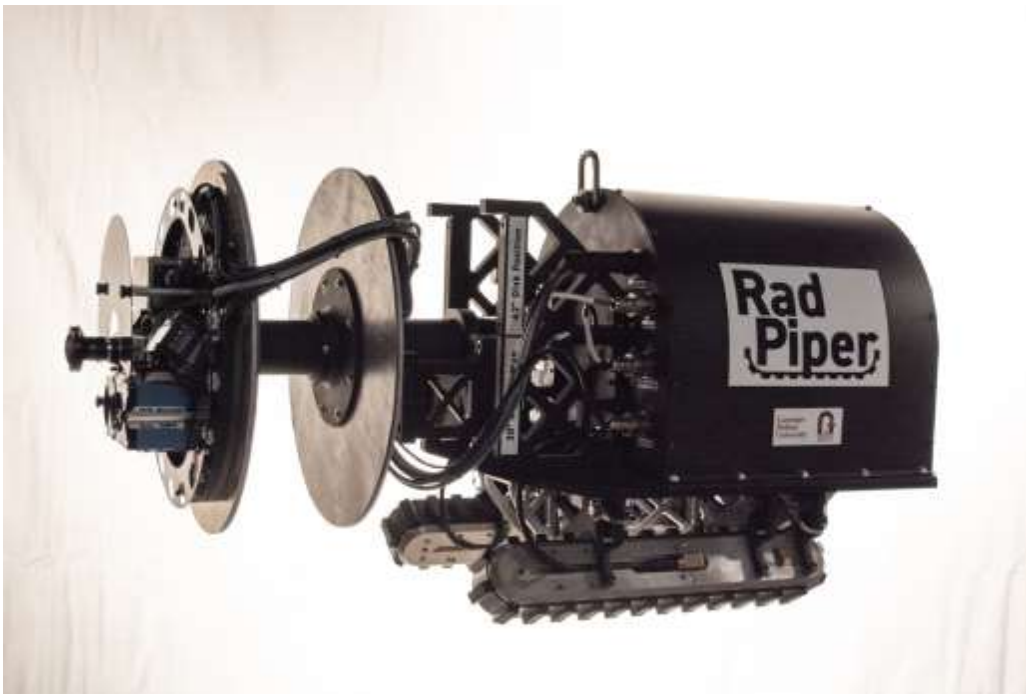

Fig. 2: U-235 deposit measurement pipe-crawling robot RadPiper

An autonomous U-235 deposit measurement pipe-crawling robot called RadPiper [1,2,3,4] (Fig. 2) is under development. RadPiper is a battery-powered and tetherless robot which self-steers using two tracks. A detector assembly is mounted on the front to acquire radiometric data. The robot is recovered from the same pipe opening from which it is launched, hence it drives the same distance out and back, measuring the same deposits twice. This achieves redundant radiometric and odometric measurements which adds further to statistical significance.

Localization is essential for this in-pipe radiation measuring robot to report the precise location of each radiation deposit measured[5,6]. Also, since the robot will measure the same deposit twice running forward and backward, the locations of the two measurements must match up with each other precisely.

**Existing solutions and challenges**

Several prior approaches have been applied for robotic in-pipe localization in other applications.

Encoder odometry is a simple, effective and well-studied robot localization method. However, localization that entirely relies on encoder odometry is vastly insufficient for the required precision and certainty at this application. Encoding suffers from accumulated error over long travelling distance. Odometry also measures the steered path which is always longer than the straight-line pipe distance. These problems cannot be completely modelled or predicted.

Tethers are adapted by some in-pipe robots for purposes of providing power, communication, safety recovery and odometry. Martra and Tur [7] measured the length of a tether cable to acquire traveled distance information. RedZone Robotics utilizes this in their sewer robots. However, this approach is not adaptable for some in-pipe robots which are not equipped with tether for varieties of reasons. For example, as a fully-autonomous robot, RadPiper precludes tether for reasons of contamination and handling.

Some localization exploits ELF-EP communication between an in-pipe robot and sensor station set up out of the pipes [8,9]. However, in many cases such as nuclear pipes for RadPiper, it is difficult or impossible to set up sensor stations or other localization assisting equipment along a pipe.

Visual odometry is an approach for odometry of some in-pipe robots [10,11]. It is not typically very robust because its performance is highly affected by how many available valid features can be found on an inner pipe surface. In some cases such as uranium enrichment piping, only a few recognizable features can be found sparsely (Fig. 3).

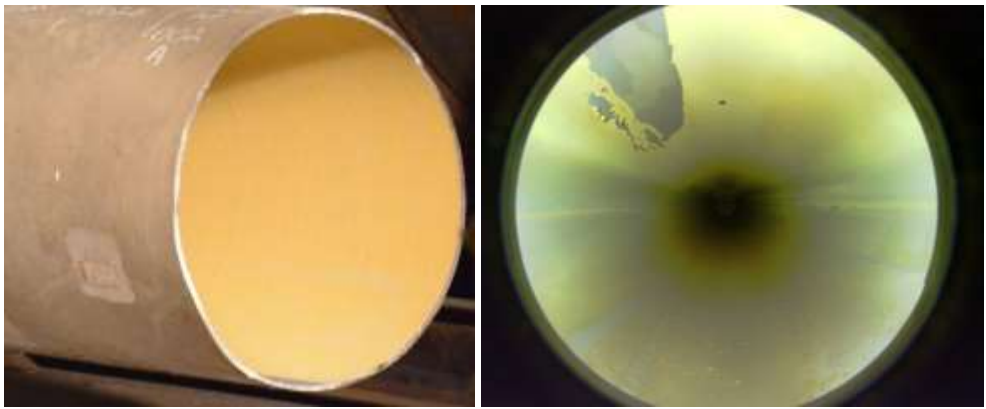

Fig. 3: No feature suitable for Visual Odometry(left); Occasional visual features are too rare and indistinct for Visual Odometry(right)

The precision of the prior localization methods is also an issue. With ELF-EP communication method [8,9], from the experiments conducted by Qi et.al, the error is about 0.75ft (0.23m) to 1.5ft (0.45m). In the research work done by Hansen et.al with visual odometry [10,11], the error was found as 0.84ft (256.4mm) when the travelling distance is about 100ft (30483mm). However, radiation measurement robots require much higher precision.

WM2019 Conference, March 3 – 7, 2019, Phoenix, Arizona, USA

**Technical approach**
The localization method developed here is based on data fusion of two sensors: laser rangefinder and track encoders.

The paper presents a reciprocal and iterative sensor data fusion technology that achieves the requisite performance.

**RECIPROCAL SENSOR FUSION ODOMETRY**
The robot RadPiper is driven by a pair of tracks. There is an encoder embedded in each track which records travelled distance. RadPiper also incorporates a laser rangefinder which measures the absolute distance. The rangefinder is intended to measure the distance between the robot and the reflecting surface at the pipe end. The whole setup is shown in Fig. 4.

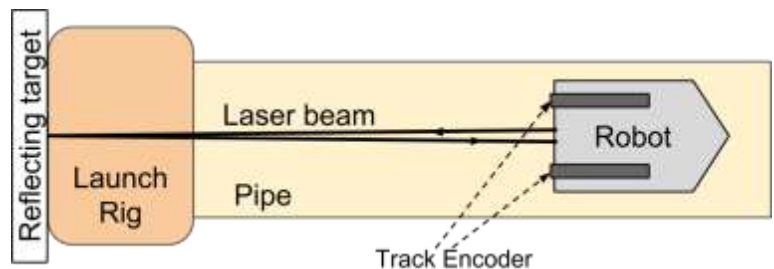

Fig. 4: Laser ranging and track encoding for localization

In this section, the paper will respectively discuss problems of raw data acquired by each sensor in and present how the problems of one sensor can be improved by the other one iteratively and reciprocally. With the improved data, the paper will then discuss how data fusion estimates the robot's trajectory. The general overall work flow of this localization method is shown in Fig. 5.

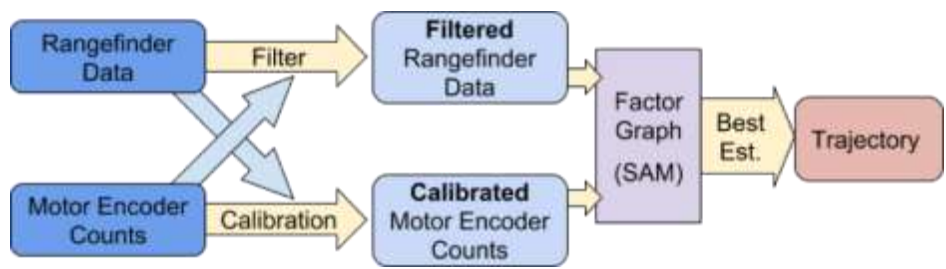

Fig. 5: Work flow for high precision in-pipe Localization with Reciprocal Sensor Fusion

**Rangefinder data filter**

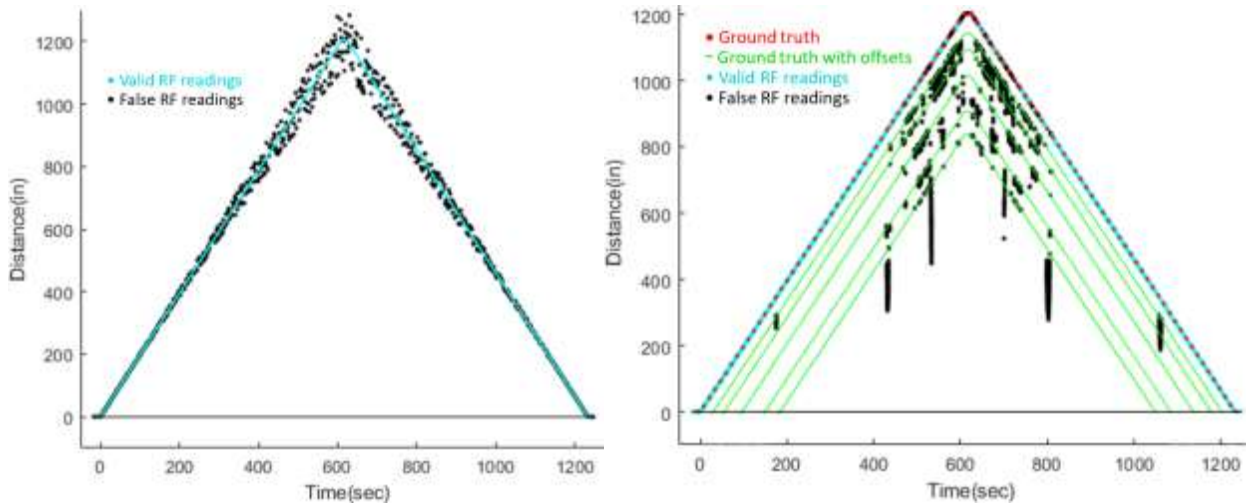

Fig. 6(a) Distance-time plot of laser Rangefinder readings with typical Gaussian noise (left).

Fig. 6(b) Distance-time plot of laser Rangefinder readings acquired while driving out and back in a pipe having 30-inch diameter and 1210-inch length during 1235 seconds (right)

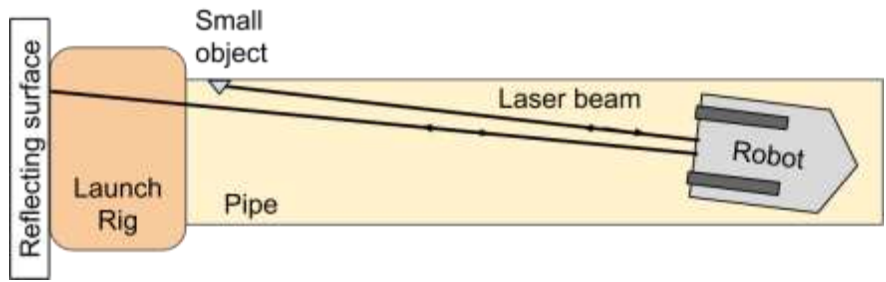

Fig. 6(c): Rangefinder false reading

A laser rangefinder is installed on RadPiper to measure the distance back to the end of a pipe. However, rangefinders report many false readings in pipes that are a consequence of striking pipe walls. The false readings are not just valid reading with random noise (Fig. 6a), but have certain patterns that are challenging for traditional filter methods.

Real rangefinder data from one test run is plotted in Fig. 6b. The horizontal axis is time in second, and the vertical axis is distance in foot. The V-shape red line is the actual distance between the robot and the end of pipe. It increases and then decreases, indicating the robot drives into the pipe and traverse back to the starting point.

The aqua and black dots on the plot are measurements from the rangefinder. Aqua data points are valid readings which lie right on the red line representing ground truth. Black data points are false readings which do not correspond to the real travelling distance of the robot. The false readings are not entirely random. Most of them can be found on the green lines, each of which is the correct trajectory being shifted by a certain distance. False measurements are caused by laser beam reflecting off pipe walls or other objects in a pipe instead of the intended surface behind the launch rig, as demonstrated in Fig. 6c.

The small object on the diagram can be a deposit lump or convex part of unsmooth pipe interior surface. False readings measure the distance from the robot to this kind of objects, which are always shorter than the actual distance, so in Fig. 6a almost all black false data points fall below the red line.

The measurements are almost all aqua valid at shorter distance within 70ft. That 70ft is a distance before the laser strikes the pipe wall at this pipe diameter. As the distance increases, less and less aqua readings but more and more black false ones appear. It is because that at long distance the rangefinder's view angle is narrowed due to the nature of the shape of pipes. Therefore, the chance of false readings happening increases as the robot drives deeper into the pipe. Though at long distance valid data points in aqua become very few, whenever there are valid measurements it is always right on red line reflecting the distance precisely.

Overall, laser rangefinder is a good sensor for localization because it is highly precise if the measurement is valid. However, at long distance valid measurements becoming very sparse and intermittent. Furthermore, false readings from rangefinder at long distance are not any typical noise. Other than offsetting, they are very similar to the correct readings, which makes it very difficult to filter.

To filter out the false readings, different approaches were developed. If a line can be found that is close enough to the trajectory, a region with a carefully chosen width then can be set around this line with boundaries above and below. Readings out of this region will be considered false readings to be eliminated. In order to find a good estimating line, line fitting methods like RANSAC[12] or voting scheme Hough transform[13] were adapted, but they did not perform well because as the robot steers forward, its travelling path is not straight. Thus, the robot's straight-line pipe distance does not increase linearly. Therefore, it is nearly impossible find a line that can work well for filtering.

One of the reasons why methods above would fail is that they try to solve the problem by fusing raw rangefinder data. To filter well, it is necessary to have a good estimate of the real trajectory. However, unfiltered raw rangefinder data includes both correct and false readings from multiple lines of ambiguous trajectories.

In the method that is being presented, encoder odometry is used for this purpose being a trajectory estimate to filter rangefinder data. Though encoder odometry has the drifting issue over long distances, in this method, this issue is avoided by iteration, which allows us to only look at a small distance for encoder odometry at any given time.

After syncing different sensors' measurements, encoder counts, and rangefinder readings share the same time stamps of measurements. Rangefinder readings are noted as $Rf[t]$ and encoder counts are noted as $Ec[t]$, while $t \in T$, $T$ is a set of discrete time stamps. During filtering, $Loc_{est}[k]$ as estimated location will also be generated at the same time as a reference value to help identify false rangefinder reading. The strategy of filtering rangefinder reading with encoder counts is presented in an iterative manner as follows:

- For $t = 0$, $Loc_{est}[0] \leftarrow 0$.
- For $t = k (k \geq 1, k \in N)$, $Loc_{est}[k]$ is obtained:
    1) Estimate $Loc_{est}[k]$ with encoder counts:
    $$Loc_{est}[k] = Loc_{est}[k-1] + \frac{Ec[k] - Ec[k-1]}{C}$$
    Here $C$ is a coefficient that converts encoder counts to actual distance and has unit as *"counts/inch"*. Its value is calculated from track and motor configurations like gear ratio, track diameter, etc.
    2) Now with a reasonable estimated location value, the actual rangefinder reading at the moment of *t=k* will then be examined. If the difference between the actual rangefinder

reading and the estimated location is larger than a selected threshold, this rangefinder reading will be marked as false reading and eliminated. If the difference is within the threshold, which means the rangefinder reading is valid and reliable, $Loc_{est}[k]$ will then be updated with this rangefinder measurement. The threshold, $thres$, is an experimental value affected by the clearance between valid and false readings and the requirement on the final accuracy.

$$if\ |(Loc_{est}[k] - Rf[k])| > thres$$
$$\textbf{then}\ \ Eliminate\ Rf[k]$$
$$\textbf{else}\ Loc_{est}[k] \leftarrow Rf[k]$$

With this approach, the false rangefinder readings can be precisely marked and eliminated. After well-filtering, all rangefinder data are valid and reliable now.

**Track encoder counts**

The robot has a left and right track, and each track is equipped with an encoder. The average of the two encoder counts reflects the travelling distance of the robot center:

$$Ec[t] = \frac{Ec_{Left}[t] + Ec_{Right}[t]}{2}$$

, where $Ec_{Left}[t]$ is the encoder counts from left track and $Ec_{Right}[t]$ is from the right track.

$$Ec[t] \propto Loc[t]$$

, where $Loc[t]$ is the robot's travelling distance, also the location when taking start point as $0$.

As discussed earlier, encoder odometry is not reliable over a long distance due to drifting issues. Drifting may be caused by slippage, robot steering, robot climbing over lumps or other accidental situations. However, after filtering rangefinder data, it becomes possible to calibrate encoder odometry using valid rangefinder reading as landmarks.

Encoder odometry is forced to meet its corresponding rangefinder reading periodically to prevent error accumulation. A step distance is set, so that whenever location is increased or decreased, by the step distance, the encoder odometry is multiplied with a coefficient to equal to the rangefinder reading taken at the same time. If the rangefinder reading was marked as a false reading at the current time stamp, the next encoder odometry data point will then be considered instead in the same way. The detailed implementation is presented below. Only the first half of the trajectory, the part of robot driving forward, is presented here.

- Initially, $Ec[0]$ is considered as "calibrated". Therefore, the flag $j$ for "the last calibrated encoder reading" is set as $0$. Searching index $k$ is set as $0$.

- Loop this section until the end of the first half of the trajectory:

    1) Keep increasing index $k$,
       until $Rf[k] - Rf[j] > Dist_{step}$,
       $Dist_{step}$ is the step distance;

    2) $Ec[j:k] \leftarrow Ec[j:k] * \frac{Rf(k) - Ec(k)/C}{Rf(k) - Rf(j)}$

    3) $j \leftarrow k$

Here, $C$ again is a coefficient in the unit of *"counts/inch"*.

In our practice, the $Dist_{step}$ is set as 3 feet.

After the operation above, any encoder odometry drifting problem is restricted within each 3-foot section. Encoder odometry is excellent over such a short distance. At this point, encoder odometry is well-calibrated by filtered rangefinder data, and ready for the following steps.

**Information fusion - generating trajectory**
In reality the two processes, rangefinder data filtering and encoder odometry calibration, happen concurrently.

With all the data from rangefinder and encoder, a factor graph can be built for state estimation to generate the robot's trajectory. In this process, GTSAM(Georgia Tech Smoothing and Mapping library)[14] was adapted. In the factor graph, position estimates are inserted as nodes and measurements as edges. Edges for encoder odometry were inserted with larger variances and edges for rangefinder data with smaller variances, because rangefinder gives absolute and accurate measurement whenever it is available.

A trajectory will be formed in the end by optimizing each state in the graph to reach the maximum likelihood.

**TESTING RESULT**
In order to evaluate this localization method, two tests were conducted with seven 100ft-long test runs in a 30-inch diameter pipe.

**Test I: Ground truth comparison**
Results from reciprocal sensor fusion odometry is compared with the ground truth which is measured by a total station survey instrument. This measures distance using a modulated infrared signal emitted by itself and reflected by a prism installed on the robot[15]. In our case, a prism is installed on the robot for total station to measure and offer ground truth. The whole setup for testing is shown in Fig. 7.

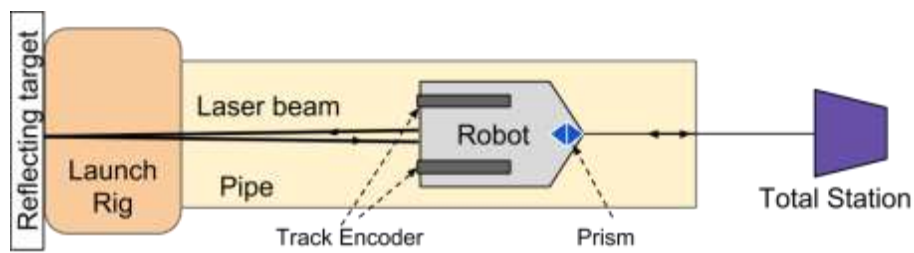

Fig. 7: Total station setup for acquiring ground truth

Error in this test is considered as the absolute value of the difference between reciprocal sensor fusion localization result and ground truth measured by total station.

$$E_1(t) = |Loc(t) - Gt(t)|$$

, $E_1(t)$ is the error, $Loc(t)$ is the localization result, and $Gt(t)$ is the ground truth. Error $E(t)$ from one random-selected test run is plotted in Fig. 8. The Error-Time plot shows that error is below 0.3 inch almost all time, and only very few extreme error values reached 0.5 inch. The statistical analysis of the ground truth comparison error from seven test runs is shown in Table I.

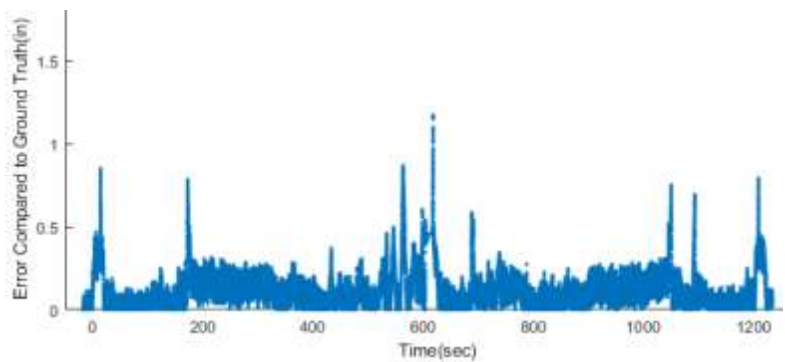

Fig. 8: Error of Reciprocal Sensor Fusion Localization vs. Ground Truth from one Test Run

TABLE I: Error (inch) of Reciprocal Sensor Fusion Localization vs. Ground Truth

| Test Run | Max($E_1$) | Mean($E_1$) | Var($E_1$) | Std($E_1$) |
|---|---|---|---|---|
| 1 | 0.96 | 0.1 | 0.0107 | 0.10 |
| 2 | 1.07 | 0.11 | 0.0155 | 0.12 |
| 3 | 0.99 | 0.12 | 0.0245 | 0.16 |
| 4 | 0.94 | 0.12 | 0.0125 | 0.11 |
| 5 | 1.17 | 0.11 | 0.0116 | 0.11 |
| 6 | 1.14 | 0.14 | 0.0259 | 0.16 |
| 7 | 1.59 | 0.13 | 0.0276 | 0.17 |
| **Max.** | **1.59** | **0.14** | **0.0276** | **0.17** |
| **Ave.** | **1.12** | **0.12** | **0.0183** | **0.13** |

**Test II: Block test**
Any in-pipe feature or uranium deposit will be detected twice on the robot's forward and backward running and it is very important that the two measured locations match up with each other precisely. This research refers to this matchup as "Zippering".

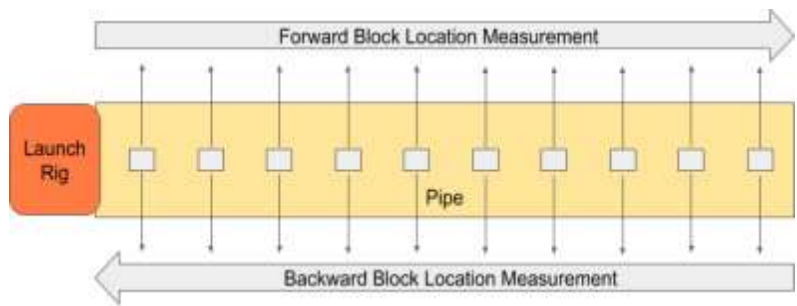

Fig. 9: Block test setup illustration

Therefore, a test was setup as shown in Fig. 9. In a 100-feet-long pipe, 25 wooden blocks were placed 4 feet apart from each other as "simulated deposits". The robot is deployed to run through the pipe and come back to the launch rig. Time stamps are recorded at the same time when blocks are detected by the Lidar installed on the robot. One block has two corresponding time stamps because the robot detects it once driving forward and again driving backward. After post-processing, based on the blocks' time stamps, the corresponding locations can be found from the generated trajectory. "Zippering error" is the difference between a block location as measured driving forward and backward.

$$E_2(n) = Block(n).Loc\left(t_f(n)\right) - Block(n).Loc\left(t_{b(n)}\right)$$

, where $n$ is the block number, $Block(n)$ indicates the n-th block, $E_2(n)$ is the "Zippering error" of the n-th block, $t_f(n)$ is the time stamp recorded when the robot detects this block during forward running and so is $t_b(N)$ for backward, and $Loc(t)$ is the reciprocal sensor fusion localization result.

This "Zippering error" from one randomly selected test run is shown in Fig. 10, and the overall statics of all seven test runs is shown in Table.2.

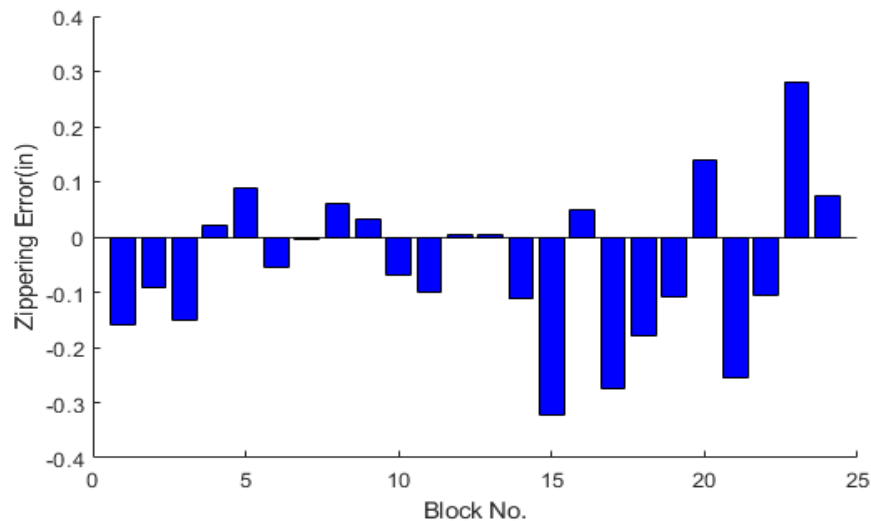

Fig. 10: "Zippering Error" from One Test Run

TABLE II: "Zippering Error" Statistics Overview

| Test Run | Max($E_1$) | Mean($E_1$) | Var($E_1$) | Std($E_1$) |
|---|---|---|---|---|
| 1 | 0.41 | 0.13 | 0.0106 | 0.10 |
| 2 | 0.46 | 0.14 | 0.0145 | 0.12 |
| 3 | 0.57 | 0.13 | 0.0178 | 0.13 |
| 4 | 0.2 | 0.08 | 0.0023 | 0.05 |
| 5 | 0.32 | 0.12 | 0.0087 | 0.09 |
| 6 | 0.36 | 0.11 | 0.01 | 0.10 |
| 7 | 0.43 | 0.13 | 0.0175 | 0.13 |
| **Max.** | **0.57** | **0.14** | **0.0178** | **0.13** |
| **Ave.** | **0.39** | **0.12** | **0.1163** | **0.10** |

**CONCLUSION**

The quality of this localization is unprecedented for any other pipe robot. This exceeds the precision and certainty required for nuclear robotic pipe Nondestructive Assay(NDA).

Reciprocal sensor fusion odometry achieves precise and certain in-pipe localization unachievable by any previous method. This particularly applies for long runs where encoder drift would otherwise be significant and absolute range sensing by laser measurement would be sparse, error-prone and intermittent.

The error between reciprocal sensor fusion localization and ground truth measurements in large diameter hundred-foot piping is typically around 0.1 inch.

The RadPiper robot localizes any feature during its forward and backward traverses. Hence, a metric of precision is the extent to which its forward and backward localization of a given feature are identical. For 24 objects over a distance of nearly 100 feet, the maximum of the "'Zippering error" is 0.4 inch, the mean error is 0.11 inch and the standard deviation of the error is 0.09 inch.

**ACKNOWLEDGEMENTS**
Funding for this work was provided by the US Department of Energy under cooperative agreement DE-EM0004383.